\documentclass[10pt,twocolumn,twoside]{IEEEtran}

\usepackage{graphicx}
\graphicspath{{data/}}

\def\etal{et al. }
\def\eg{e.g. }

\begin{document}

\title{Deep Learning Algorithms with Applications to Video Analytics for A  Smart City: A Survey}

\author{Li Wang,~\IEEEmembership{Member,~IEEE,}
        and Dennis Sng
%        Helen,~\IEEEmembership{Member,~IEEE}
\thanks{L. Wang and D. Sng are with the Rapid-Rich Object Search (ROSE) Lab at the Nanyang Technological University, Singapore 637553 (e-mail: wa0002li@e.ntu.edu.sg; dennis.sng@ntu.edu.sg).}}

%\markboth{Transaction on Image Processing,~Vol.~xx, No.~xx, xx~20xx}%
%{Shell \MakeLowercase{\textit{et al.}}: Bare Demo of IEEEtran.cls
%for Journals}

\maketitle

\begin{abstract}
Deep learning has recently achieved very promising results in a wide range of areas such as computer vision, speech recognition and natural language processing. It aims to learn hierarchical representations of data by using deep architecture models. In a smart city, a lot of data (\eg videos captured from many distributed sensors) need to be automatically processed and analyzed. In this paper, we review the deep learning algorithms applied to video analytics of smart city in terms of different research topics: object detection, object tracking, face recognition, image classification and scene labeling.
\end{abstract}


\IEEEpeerreviewmaketitle

\section{Introduction}

\IEEEPARstart{A}{}smart city aims to improve quality and performance of urban services by using digital technologies or information and communication technologies. Data analytics plays an important role in smart cities. Many sensors are installed in a smart city to capture a huge volume of data such as surveillance videos, environment and transportation data. To capture useful information from such big data, machine learning algorithms are often used and have achieved very promising results in a wide range of applications, \eg video analytics. Therefore, leveraging on machine learning can facilitate smart city development.

Machine learning aims to develop the computer algorithms which can learn experience from example inputs and make data-driven predictions on unknown test data. Such algorithms can be divided into two categories: supervised learning (\eg \cite{DBLP:journals/ml/CortesV95} \cite{rumelhart1988learning}) and unsupervised learning (\eg \cite{kohonen1982self} \cite{hotelling1933analysis}). Given labeled input and output pairs, supervised learning aims to find a mapping rule for predicting outputs of unknown inputs. In contrast, unsupervised learning focuses on exploring intrinsic characteristics of inputs. Supervised learning and unsupervised learning are complementary to each other. Since supervised learning leverages labels of inputs which are meaningful to human, it is easy to apply this kind of learning algorithms to pattern classification and data regression problems. However, supervised learning relies on labeled data which could cost a lot of manual works. Moreover, there are uncertainties and ambiguities in labels. In other words, the label for an object is not unique. To mitigate these problems, unsupervised learning can be used to handle intra-class variation as it does not require labels of data. In the past decades, machine learning methods have been applied to a wide range of applications such as bioinformatics, computer vision, medical diagnosis, natural language processing, robotics, sentiment analysis, speech recognition and stock market analysis, etc. In this paper, we focus on reviewing the recent achievements of deep learning which is a subfield of machine learning.

In contrast with shallow learning algorithms, deep learning aims to extract hierarchical representations from large-scale data (\eg images and videos) by using deep architecture models with multiple layers of non-linear transformations. With such learned feature representations, it becomes easier to achieve better performance than using raw pixel values or hand-crafted features. The principle behind this success is that deep learning is able to disentangle different levels of abstractions embedded in observed data by elaborately designing the layer depth and width, and properly selecting features that are beneficial for learning tasks. 

In fact, the history of deep learning starts at least from 1980, when Neocognitron \cite{fukushima1980neocognitron} is proposed by Fukushima. In 1989, LeCun \etal \cite{DBLP:journals/neco/LeCunBDHHHJ89} propose to apply backpropagation onto a deep neural network for handwritten ZIP code recognition. However, the training time on the network was too long for practical use. Also, deep neural networks have been studied in speech recognition for many years, but can hardly surpass the shallow generative models. It is due to the fact that deep learning architectures require large training data which was scarce in those early days. Hinton \etal \cite{hinton2012deep} review these difficulties and claim their confidence of solving these issues for applying deep learning to speech recognition, since Hinton \cite{hinton2007learning} has achieved breakthroughs on training multi-layer neural networks by pre-training one layer at a time as an unsupervised restricted Boltzmann machine and then using supervised backpropagation for fine-tuning. Since the breakthrough of deep learning, it has been applied to many other research areas besides speech recognition.

\begin{figure}
	\begin{center}
		\includegraphics[width=1.0\linewidth]{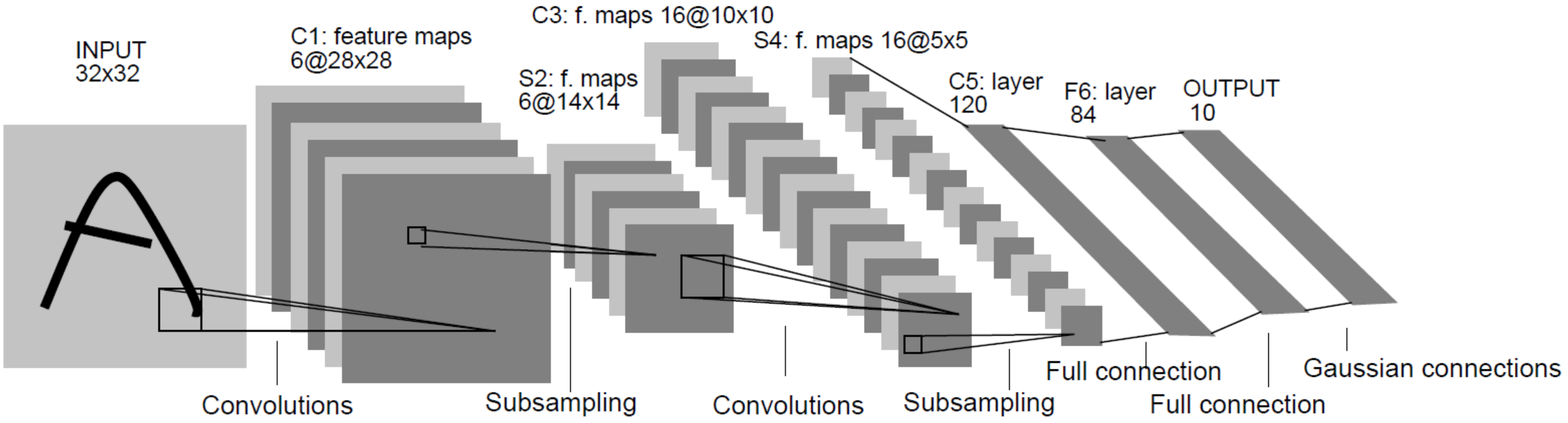}
	\end{center}
	\caption{Illustration of the CNN proposed by LeCun \etal \cite{lecun1998gradient} for digits recognition.}
	\label{fig:CNN_digitRecog}
\end{figure}

Deep learning architectures have different variants such as Deep Belief Networks (DBN) \cite{hinton2009deep}, Convolutional Neural Networks (CNN) \cite{DBLP:conf/nips/KrizhevskySH12}, Deep Boltzmann Machines (DBM) \cite{salakhutdinov2009deep} and Stacked Denoising Auto-Encoders (SDAE) \cite{vincent2010stacked}, etc. The most attractive model is Convolutional Neural Networks which have achieved very promising results in both computer vision and speech recognition. An illustration of the CNN proposed by LeCun \etal \cite{lecun1998gradient} for digits recognition is presented in Figure~\ref{fig:CNN_digitRecog}. As shown in the figure, a CNN usually consists of convolutional layers, pooling layers and fully connected layers. With loss layers on the top of the CNN, the whole network can be trained end-to-end by using the backpropagation algorithm. Compared to other deep feed-forward neural networks, a CNN is easier to train as it has fewer parameters to estimate. As a result, CNN has a wide range of applications such as image classification \cite{DBLP:journals/corr/ZuoSWLWW15}, face recognition \cite{DBLP:conf/cvpr/TaigmanYRW14}, object tracking \cite{DBLP:conf/bmvc/LiLP14}, pedestrian detection \cite{DBLP:conf/cvpr/SermanetKCL13}, attribute prediction \cite{DBLP:journals/tmm/AbdulnabiWLJ15}, scene labeling \cite{DBLP:conf/cvpr/ShuaiWZWZ15}, person re-identification \cite{DBLP:journals/corr/VariorWL14}, RGB-D object recognition \cite{DBLP:journals/tmm/0001LCCW15}, image labeling \cite{DBLP:journals/spl/ShuaiZW15}, scene image classification \cite{DBLP:journals/pr/ZuoWSZY15}, speech recognition \cite{DBLP:conf/icassp/DengAY13} and natural language processing \cite{DBLP:conf/icml/CollobertW08}, etc.

Deep learning has received much attention from not only academic but also industry. For example, Geoff Hinton and Li Deng started their collaborations from 2009 in the focus of applying deep learning to large-scale speech recognition, in which the performance is significantly improved against the traditional generative models by using big training data and the correspondingly designed deep neural networks. Another exciting example is that Andrew Ng and Jeff Dean from the Google Brain team successfully extract object-level semantics (\eg cats) from unlabeled YouTube videos by using a neural network with the self-taught capacity. In future, deep learning will have more and more real applications in industry.

\begin{figure}
	\begin{center}
		\includegraphics[width=1.0\linewidth]{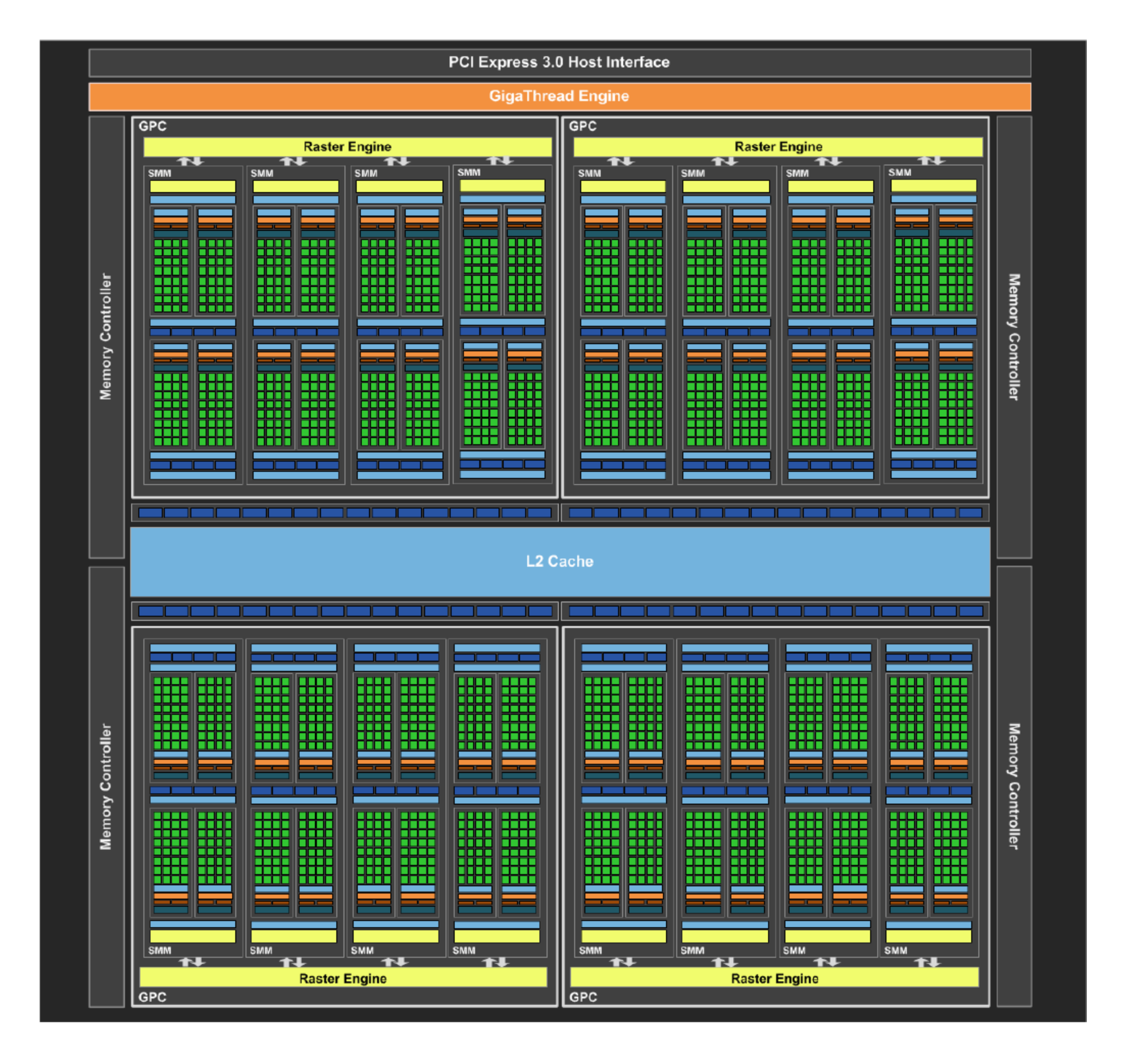}
	\end{center}
	\caption{Illustration of the parallel structure of the Nvidia GeForce GTX 980.}
	\label{fig:GTX980}
\end{figure}

Besides methodology breakthroughs and available big training data, the recent success for deep learning is also due to advances in hardware. Specifically, an electronic circuit called Graphics Processor Unit (GPU) is designed for accelerating the algorithms that need to process a large number of blocks of data, and is characteristic of its highly parallel structure. For example, Nvidia's latest GPU, the GTX $980$, is based on their $10$th generation GPU architecture, called Maxwell, which delivers double the performance per watt compared to the previous generation. The GM $204$ chip is composed of an array of $4$ Graphics Processing Clusters (GPCs), $16$ Streaming Multiprocessors (SMs), and $4$ memory controllers. The GeForce GTX $980$ uses the full complement of these architectural components. Its parallel structure is illustrated in Figure~\ref{fig:GTX980}. It is necessary to mention that the Nvidia company has made a lot of contributions for popularizing GPUs, \eg Nvidia marketed the GeForce 256 as the world's first GPU. Recently, GPUs are very popular in machine learning. For a general instance, Raina \etal \cite{DBLP:conf/icml/RainaMN09} propose to accelerate the sparse coding algorithm \cite{DBLP:conf/nips/LeeBRN06} by using GPUs and finally speed up the previous method using a dual-core CPU up to $15$ times. In particular, Raina \etal \cite{DBLP:conf/icml/RainaMN09} also report that the GPU speedup on learning Deep Belief Networks (DBNs) achieves up to $70$ times against a dual-core CPU implementation. For learning a four-layer DBN with $100$ million parameters, using GPU rather than CPU is able to reduce the required time from several weeks to around a single day.

Smart city is so-called ``smart" as it has the capability of computing and analyzing urban data collected from \eg monitoring systems, government agencies, commercial companies and social networking websites. Since deep learning is suitable for handling large-scale data, it can be used to process and analyze millions of video data captured from the distributed sensors in a smart city. Regarding such data, there are many active research topics such as object detection, object tracking, face recognition, image classification and scene labeling. In the following sections, we review the state-of-the-art deep learning algorithms in these application areas.

\section{Object Detection}
Object detection aims to precisely locate interested objects in video frames. Many interesting works have been proposed for object detection by using deep learning algorithms. We review some representative works as follows. 

\begin{figure}
	\begin{center}
		\includegraphics[width=1.0\linewidth]{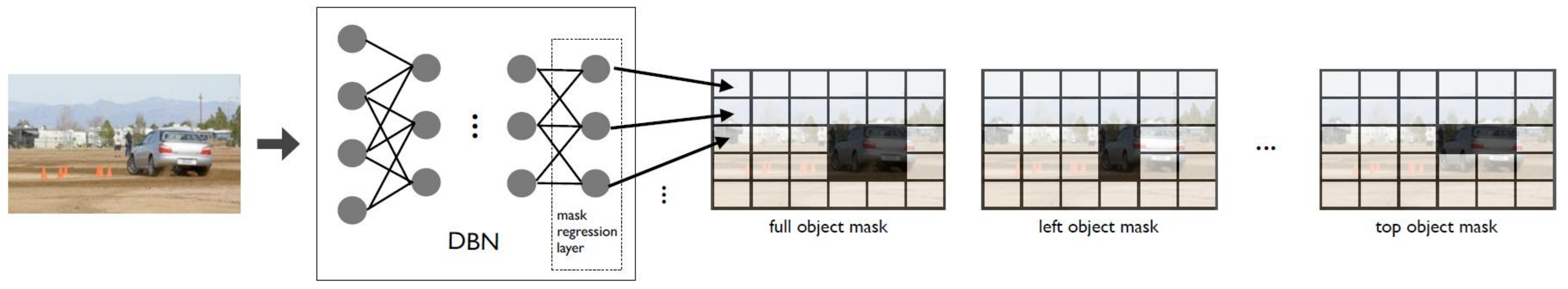}
	\end{center}
	\caption{Illustration of object detection as DNN-based regression proposed by Szegedy \etal \cite{DBLP:conf/nips/SzegedyTE13}.}
	\label{fig:DNN_ObjDet_NIPS2013}
\end{figure}

\begin{figure}
	\begin{center}
		\includegraphics[width=1.0\linewidth]{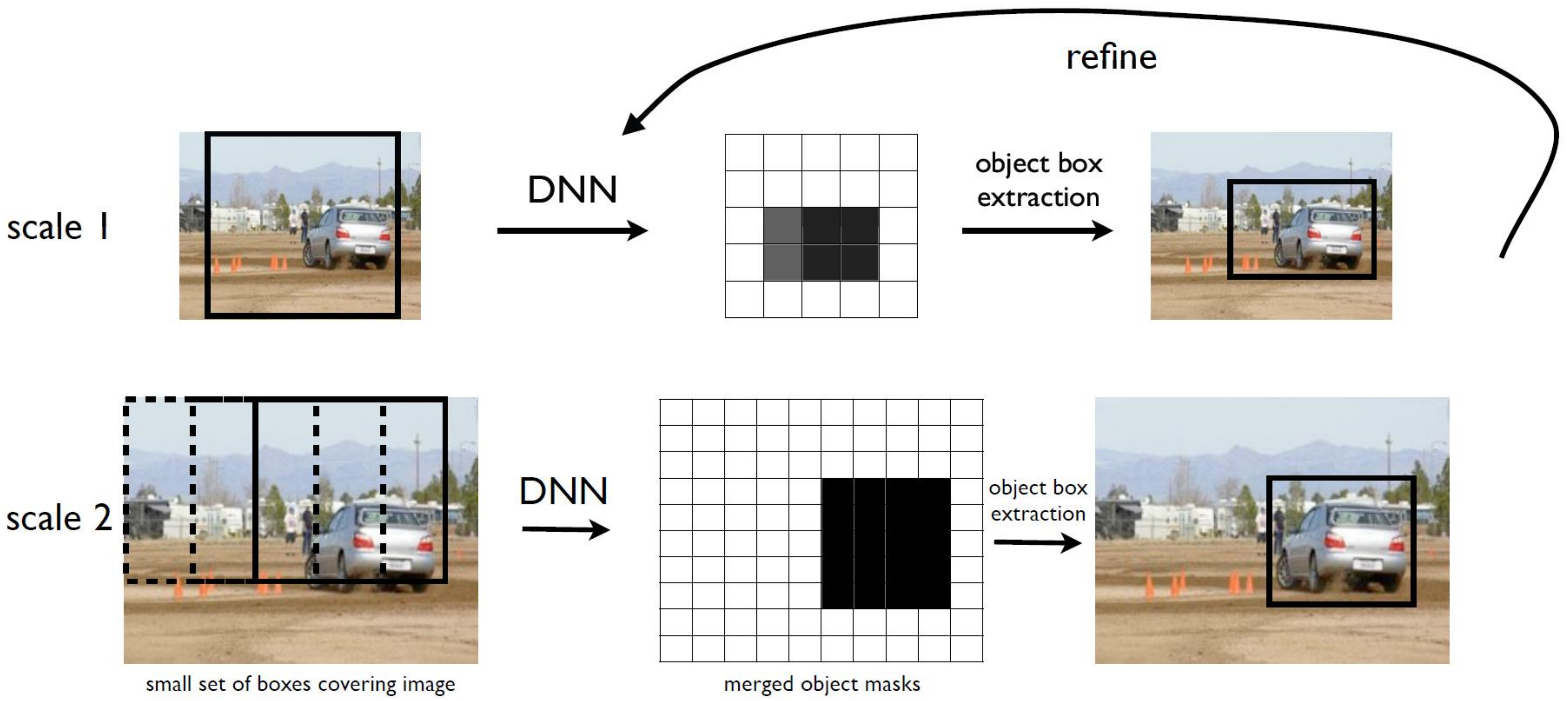}
	\end{center}
	\caption{Illustration of multi-scale strategy for refining detection precision proposed by Szegedy \etal \cite{DBLP:conf/nips/SzegedyTE13}.}
	\label{fig:DNN_ObjDet_NIPS2013_2}
\end{figure}

Szegedy \etal \cite{DBLP:conf/nips/SzegedyTE13} modify deep convolutional neural networks \cite{DBLP:conf/nips/KrizhevskySH12} by replacing the last layer with a regression layer to produce a binary mask of the object bounding box as illustrated in Figure~\ref{fig:DNN_ObjDet_NIPS2013}. Moreover, a multi-scale strategy (see Figure~\ref{fig:DNN_ObjDet_NIPS2013_2}) is proposed for the DNN mask generation to refine detection precision. As a result, the average precision $0.305$ over $20$ classes on Pascal Visual Object Challenge (VOC) $2007$ can be achieved by applying the proposed network a few dozen times per input image.

\begin{figure}
	\begin{center}
		\includegraphics[width=1.0\linewidth]{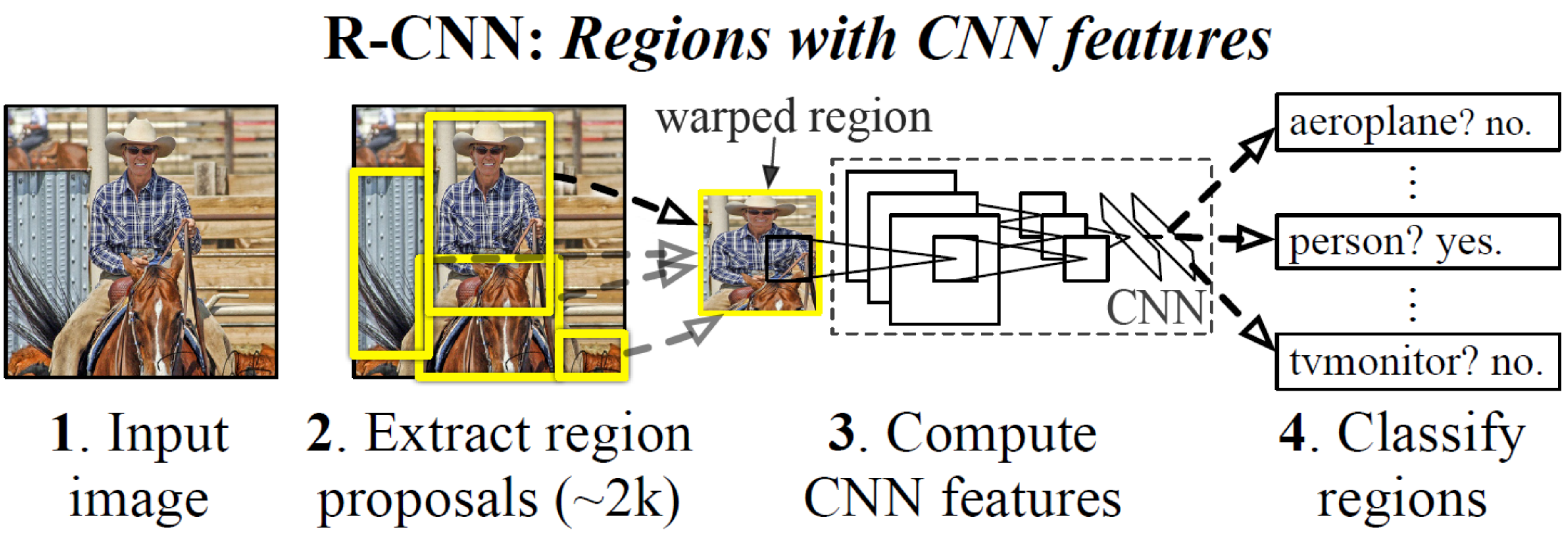}
	\end{center}
	\caption{Illustration of R-CNN: Regions with CNN features proposed by Girshick \etal \cite{DBLP:conf/cvpr/GirshickDDM14}.}
	\label{fig:R-CNN}
\end{figure}

Different from the Szegedy's method \cite{DBLP:conf/nips/SzegedyTE13}, Girshick \etal \cite{DBLP:conf/cvpr/GirshickDDM14} propose a bottom-up region proposal based deep model for object detection. Figure~\ref{fig:R-CNN} presents an overview of the method. First, around $2000$ region proposals are generated within the input image. For each proposal, a large convolutional neural network is used to extract features. Finally, each region is classified by using class-specific linear SVMs. It is reported that using this method can significantly improve detection performance on $VOC~2012$ by more than $30 \%$ in terms of mean average precision (mAP).

Similarly, Erhan \etal \cite{DBLP:conf/cvpr/ErhanSTA14} propose a saliency-inspired deep neural networks to detect any object of interest. However, a small number of bounding boxes are generated as object candidates by using a deep neural network in a class agnostic manner. In this work, object detection is defined as a regression problem to the coordinates of bounding boxes. For the training part, an assignment problem between predictions and groundtruth boxes is solved to update box coordinates, their confidences and the learned features by using backpropagation. In summary, a deep neural network is customized towards object localization problem.

Object detection has a wide range of smart city applications, such as pedestrian detection, on-road vehicle detection, unattended object detection. Deep learning algorithms are able to handle large variations of different objects. As a result, the smart city systems using deep models are more robust to large-scale real data.

\section{Object Tracking}
Object tracking is intended to locate a target object in a video sequence given its location in the first frame. Recently, some deep learning based tracking algorithms have achieved very promising results. We review some representative works as follows.

\begin{figure}
	\begin{center}
		\includegraphics[width=1.0\linewidth]{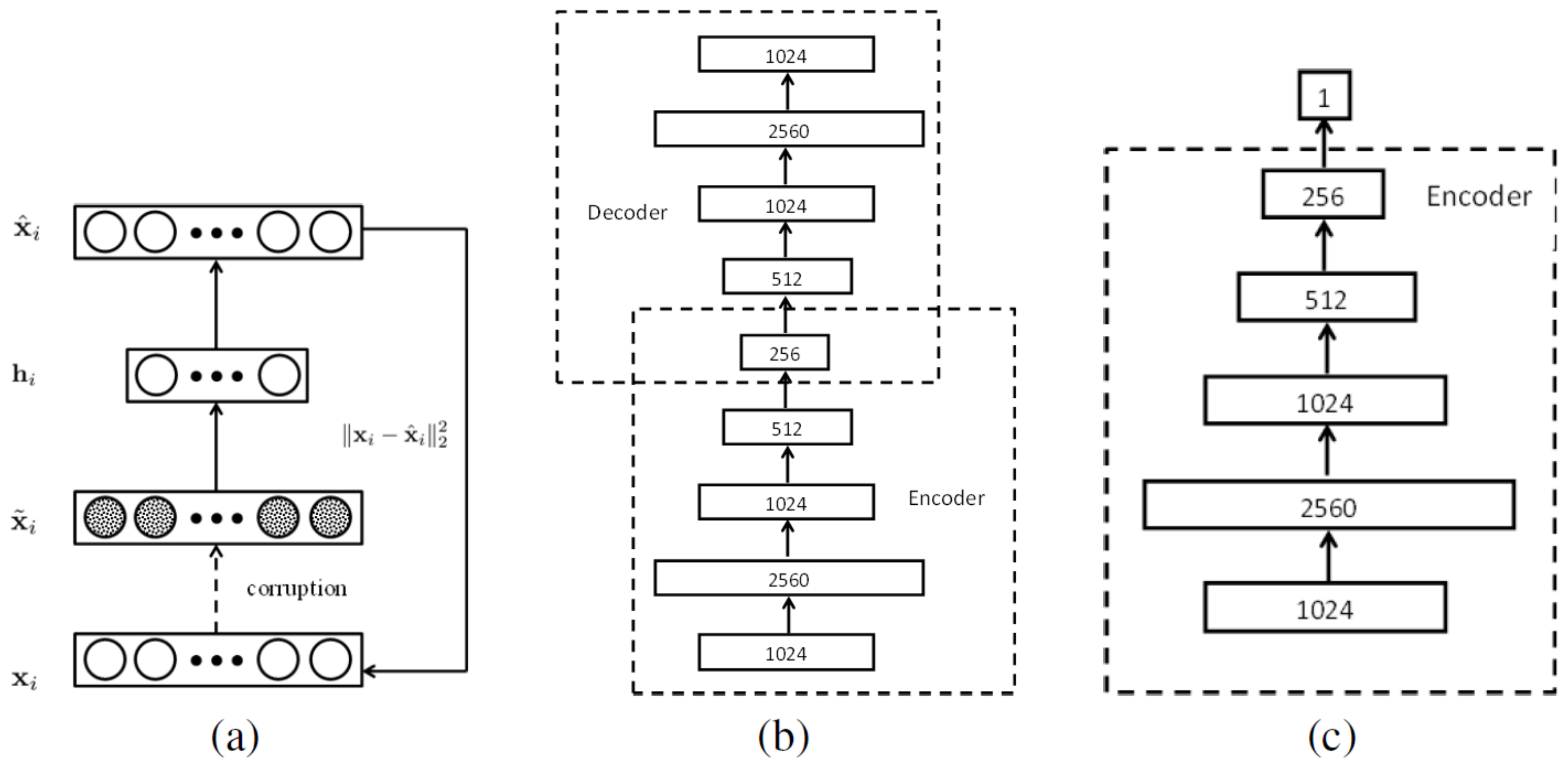}
	\end{center}
	\caption{Illustration of the network architecture proposed by Wang and Yeung \cite{DBLP:conf/cvpr/GirshickDDM14}: (a) denoising autoencoder; (b) stacked denoising autoencoder; (c) network for online tracking.}
	\label{fig:SDAE}
\end{figure}

Wang and Yeung \cite{DBLP:conf/nips/WangY13} propose to learn deep and compact features for visual tracking by using stacked denoising autoencoder \cite{DBLP:journals/jmlr/VincentLLBM10}. The network architecture is illustrated in Figure~\ref{fig:SDAE}. It is reported that using the learned deep features can outperform other $7$ state-of-the-art trackers on $10$ sequences in terms of two tracking measurements: average center error ($7.3$ pixels) and average success rate ($85.5$\%). Additionally, the proposed tracker with deep features achieved an average frame rate of $15$fps which is suitable for real-time applications.

\begin{figure}
	\begin{center}
		\includegraphics[width=1.0\linewidth]{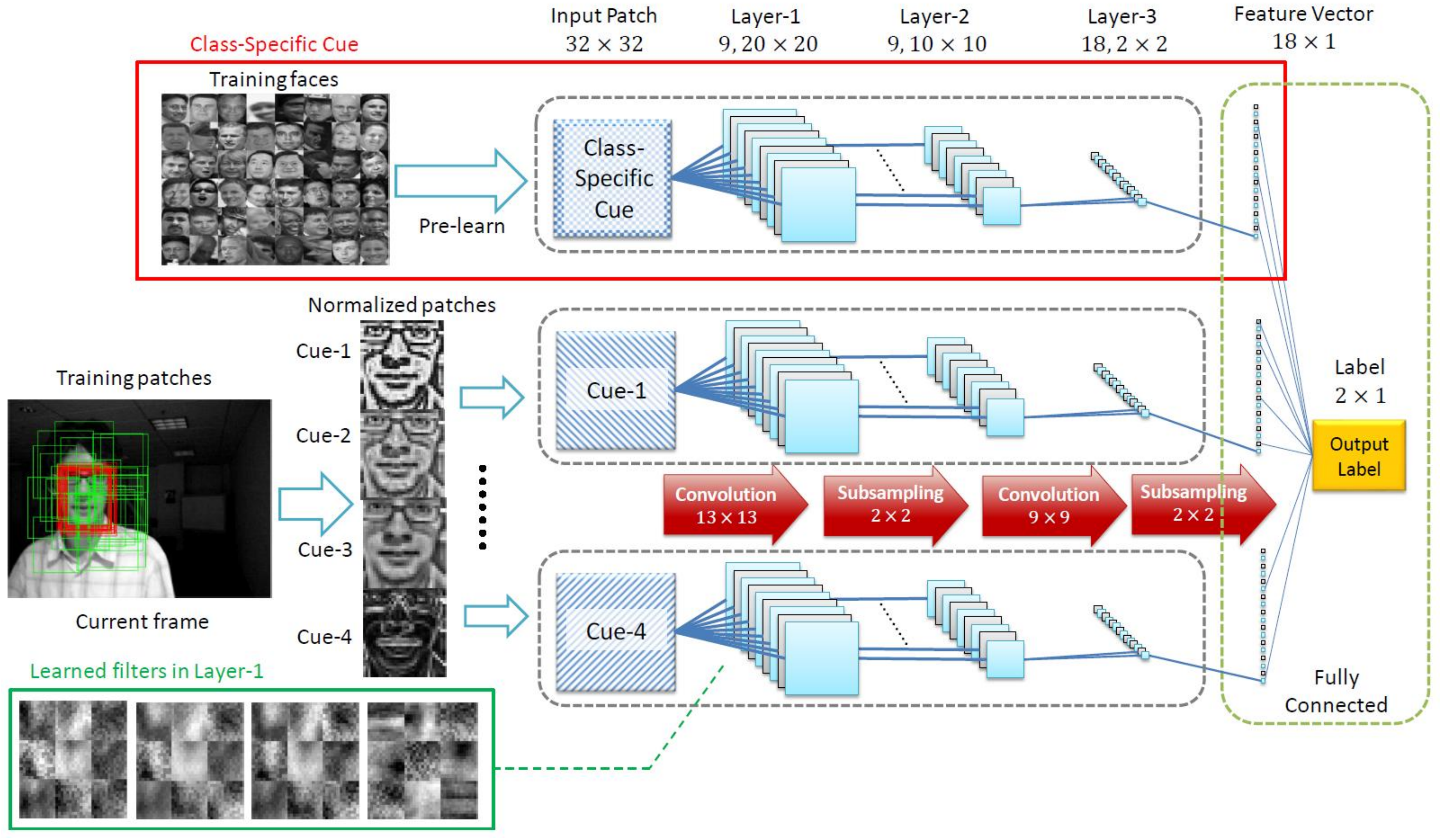}
	\end{center}
	\caption{Overview of the network architecture proposed by Li \etal \cite{DBLP:conf/bmvc/LiLP14}.}
	\label{fig:DeepTrack}
\end{figure}

Li \etal \cite{DBLP:conf/bmvc/LiLP14} propose to learn discriminative feature representations for visual tracking by using convolutional neural networks. An overview of the method is illustrated in Figure~\ref{fig:DeepTrack}. It can be observed that a pool of multiple convolutional neural networks are utilized to maintain different kernels regarding all possible low-level cues, which aim to discriminate object patches from their surrounding background. Given a frame, the most prospective convolutional neural network in the pool is used to predict the new location of the target object. Meanwhile, the selected network is retrained using a warm-start backpropagation scheme. Also, we can observe from Figure~\ref{fig:DeepTrack} that a class-specific convolutional neural network is involved in the whole architecture, which is helpful for tracking a certain class of objects (\eg a person's face).

\begin{figure}
	\begin{center}
		\includegraphics[width=1.0\linewidth]{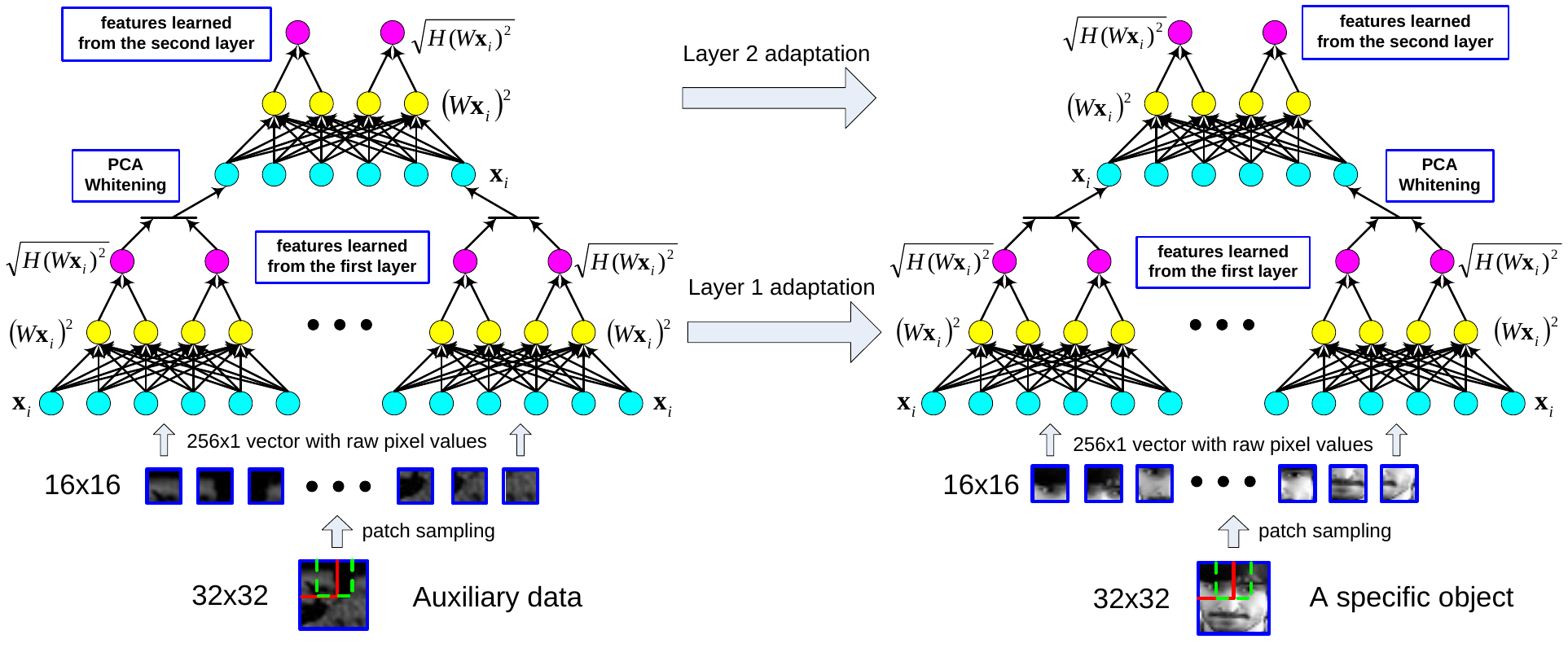}
	\end{center}
	\caption{Illustration of the stacked architecture and the adaptation module proposed by Wang \etal \cite{DBLP:journals/tip/WangLWCY15}.}
	\label{fig:OurTIP15}
\end{figure}

Wang \etal \cite{DBLP:journals/tip/WangLWCY15} propose to learn hierarchical features robust to both complicated motion transformations and target appearance changes. The stacked architecture and the adaptation module are illustrated in Figure~\ref{fig:OurTIP15}. First, the generic features robust to complicated motion transformations are offline learned from auxiliary video data by using a two-layer neural networks under the temporal constraints \cite{DBLP:conf/nips/ZouNZY12}. Then, the pre-learned features are online adapted according to the specific target object sequence. As a result, the adapted features are able to capture appearance changes of target objects. For example, the proposed tracker can handle not only non-rigid deformation of a basketball player's body but also the specific appearance changes. It is reported that using the feature learning algorithm can significantly improve tracking performance, especially on the sequences with complex motion transformations.

Object tracking can be applied to surveillance systems of smart cities. It is important to automatically track suspected people or target vehicles for safety monitoring and urban flow management. Deep learning can leverage smart city big data to train deep models which are more robust to visual variations of target objects than traditional models. Therefore, smart city tracking systems can be enhanced by using deep learning algorithms for handling large amount of video data. 

\section{Face Recognition}
Face recognition consists of two main tasks: face verification and face identification. The former aims to determine whether the given two faces belong to the same person. The latter is intended to find the identities of the given faces from the known face set. Recently, many deep learning based algorithms have achieved very promising results in these two face recognition tasks. We review some of them as follows.

\begin{figure}
	\begin{center}
		\includegraphics[width=1.0\linewidth]{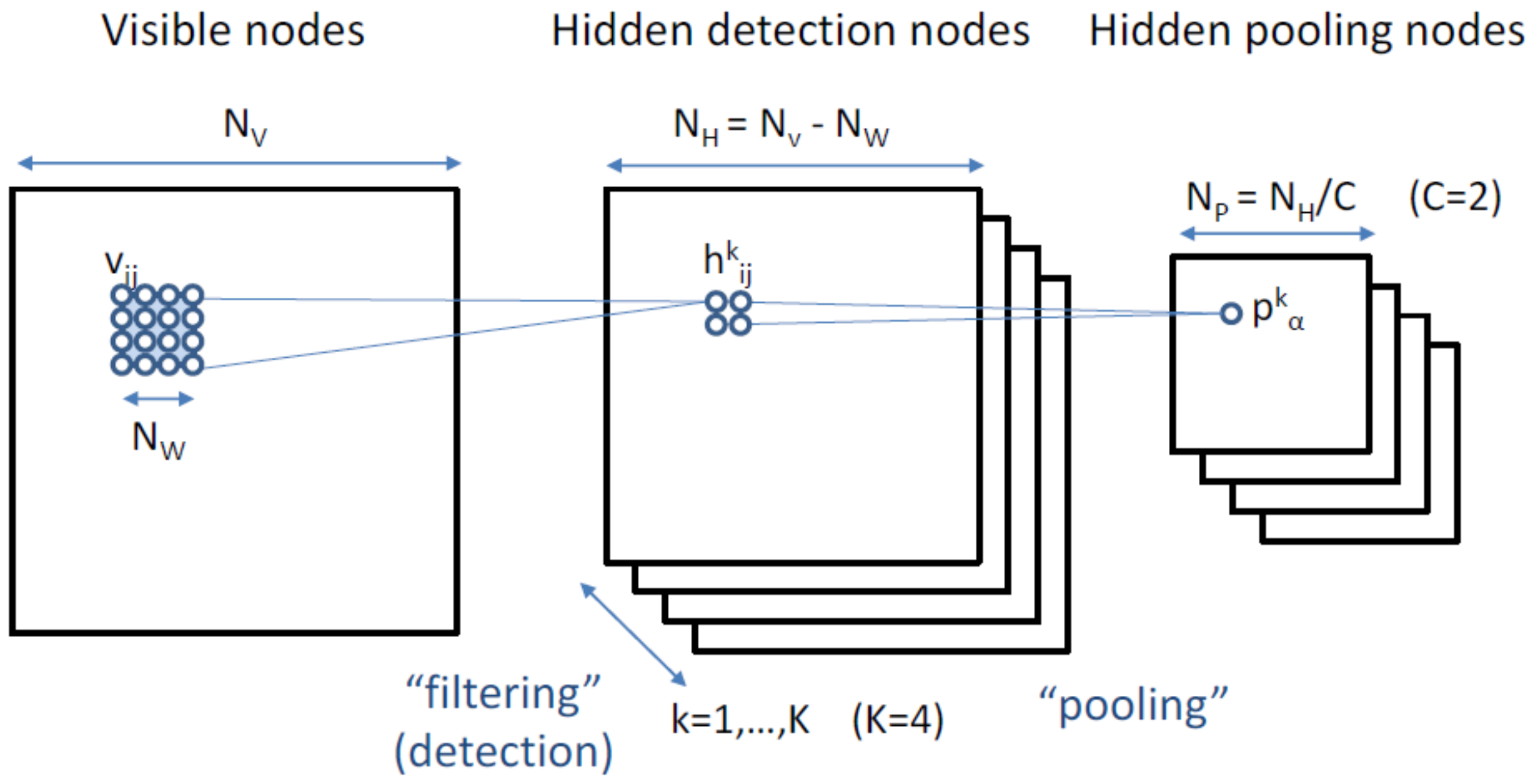}
	\end{center}
	\caption{Illustration of the convolutional restricted Boltzmann machine used in Huang \etal \cite{DBLP:conf/cvpr/HuangLL12}.}
	\label{fig:CRBM}
\end{figure}

Huang \etal \cite{DBLP:conf/cvpr/HuangLL12} propose to learn hierarchical features for face verification by using convolutional deep belief networks. The main contributions of this work are as follows: i) a local convolutional restricted Boltzmann machine is developed to adapt to the global structure in an object class (\eg face); ii) deep learning is applied to local binary pattern representation \cite{DBLP:journals/pr/OjalaPH96} rather than raw pixel values to capture more complex characteristics of hand-crafted features; iii) learning the network architecture parameters is evaluated to be necessary for enhancing the multi-layer networks. The convolutional restricted Boltzmann machine used in the proposed method is illustrated in Figure~\ref{fig:CRBM}. It is reported that using the learned representations can achieve comparable performance with state-of-the-art methods using hand-crafted features. Actually, the subsequent works have shown that deep features outperform hand-crafted features significantly.

\begin{figure}
	\begin{center}
		\includegraphics[width=1.0\linewidth]{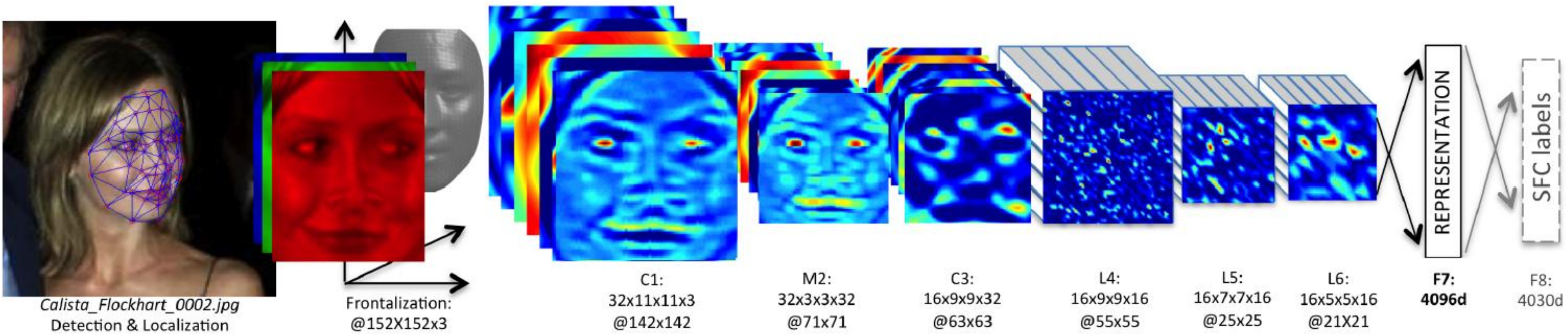}
	\end{center}
	\caption{Overview of the nine-layer deep neural network used in Taigman \etal \cite{DBLP:conf/cvpr/TaigmanYRW14}.}
	\label{fig:DeepFace}
\end{figure}

Taigman \etal \cite{DBLP:conf/cvpr/TaigmanYRW14} propose a $3$D face model based face alignment algorithm and a face representation learned from a nine-layer deep neural network. An overview of the architecture of the deep network is illustrated in Figure~\ref{fig:DeepFace}. The first three convolutional layers are used to extract low-level features (\eg edges and textures). The next three layers are locally connected to learn a different set of filters for each location of a face image since different regions have different local statistics. The top two layers are fully connected to capture correlations between features captured in different parts of a face image. At last, the output of the last layer is fed to a K-way softmax which predicts class labels. The objective of training is to maximize the probability of correct class by minimizing the cross-entropy loss for each training sample. It is shown that using the learned representations can achieve the near-human performance on the Labeled Faces in the Wild benchmark (LFW). 

\begin{figure}
	\begin{center}
		\includegraphics[width=1.0\linewidth]{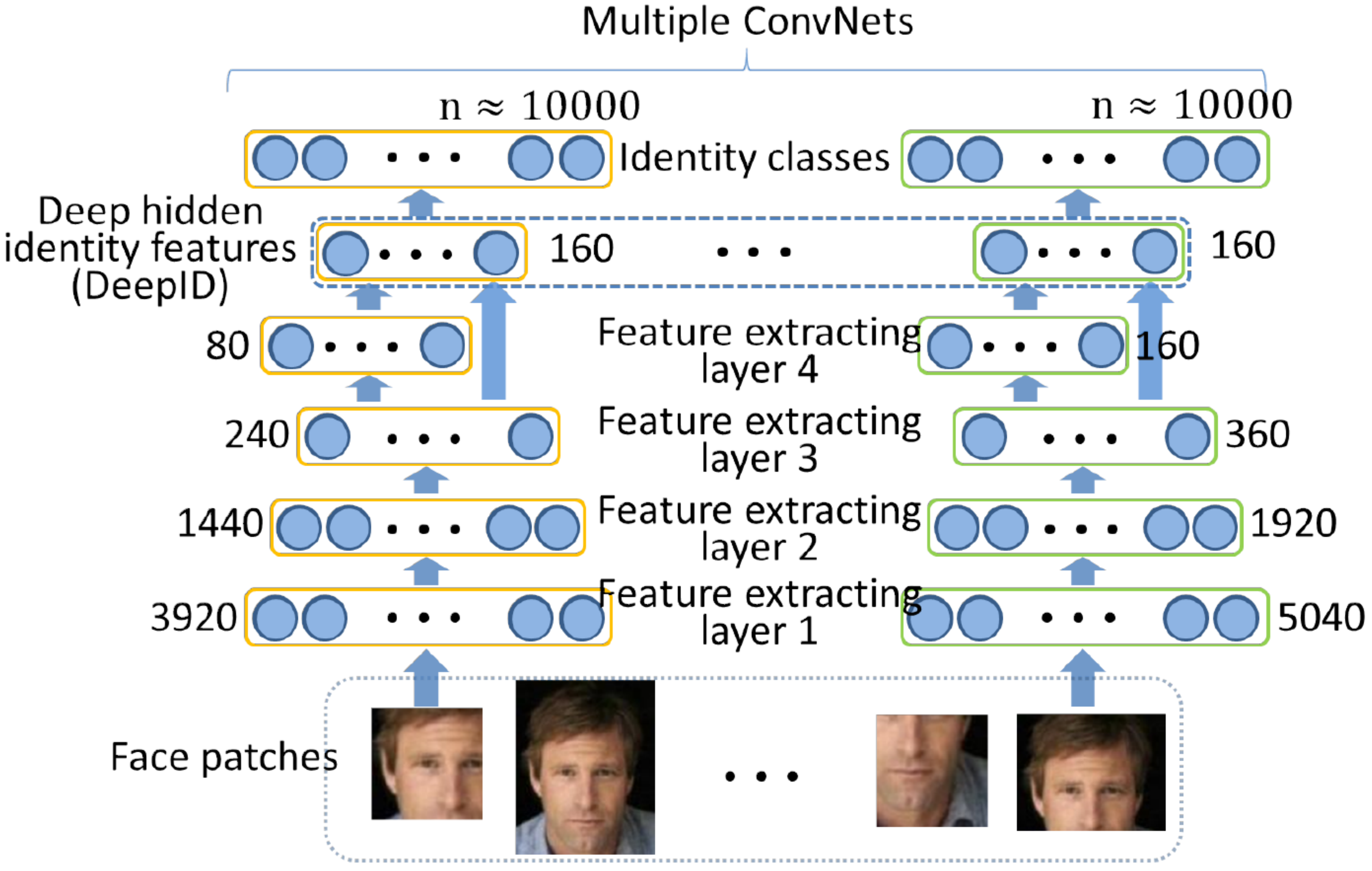}
	\end{center}
	\caption{Illustration of the feature extraction process proposed in Sun \etal \cite{DBLP:conf/cvpr/SunWT14}.}
	\label{fig:DeepID}
\end{figure}

Sun \etal \cite{DBLP:conf/cvpr/SunWT14} propose to learn so-called Deep hidden IDentity features (DeepID) for face verification. Figure~\ref{fig:DeepID} illustrates the feature extraction process. First, the local low-level features of an input face patch are extracted and fed into a ConvNet \cite{lecun1998gradient}. Then, the feature dimension gradually decreases to $160$ through several feed-forward layers, during which more global and high-level features are learned. Last, the identity class (among $10,000$ classes) of the face patch is predicted directly by using the $160$-dimensional DeepID. Rather than training a binary classifier for each face class, Sun \etal simultaneously classify all ConvNets regarding $10,000$ face identities. The advantages of this manipulation are as follows: i) effective features are extracted for face recognition by using the super learning capacity of neural networks; ii) the hidden features among all identities are shared by adding a strong regularization to ConvNets. It is reported that using the learned DeepID can achieve the near-human performance on the LFW dataset although only weakly aligned faces are used.

\begin{figure}
	\begin{center}
		\includegraphics[width=1.0\linewidth]{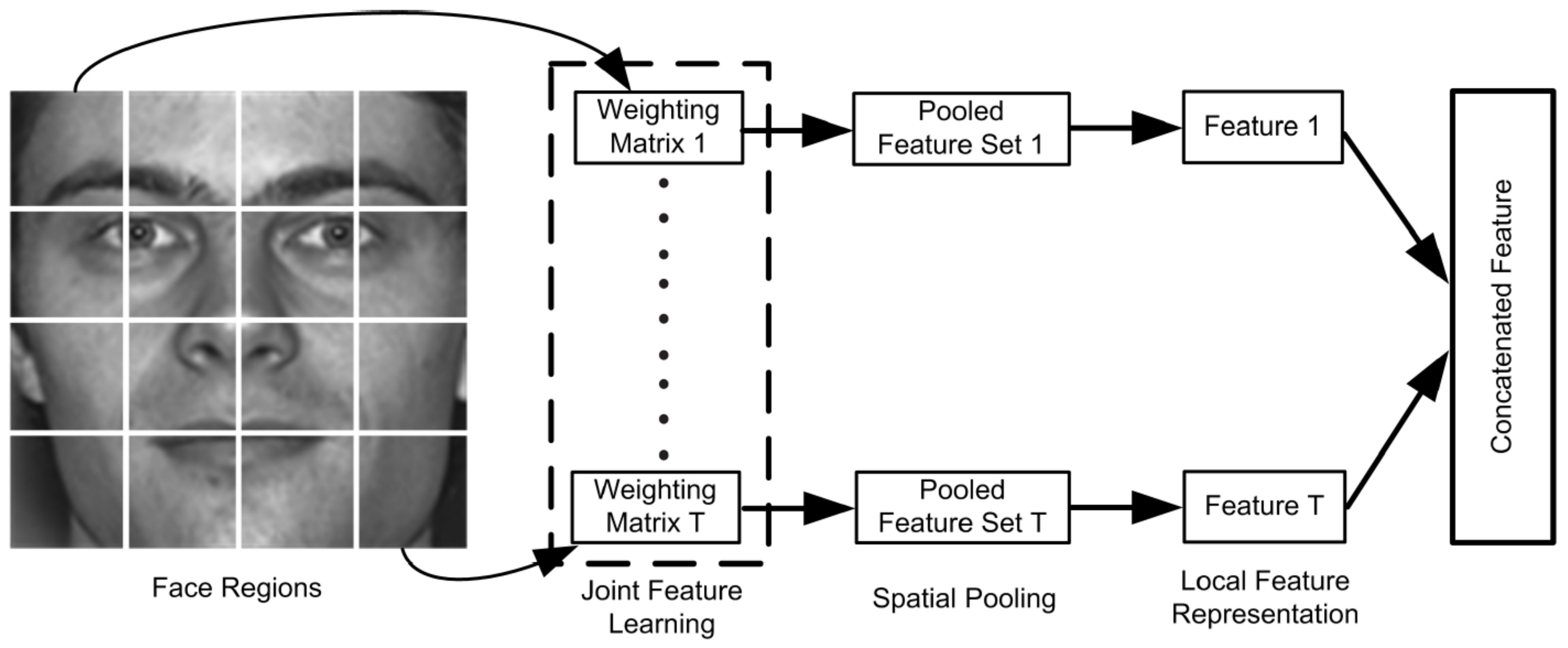}
	\end{center}
	\caption{Illustration of the basic idea of the approach \cite{DBLP:journals/tifs/LuLWM15}.}
	\label{fig:TIFS2015}
\end{figure}

Lu \etal \cite{DBLP:journals/tifs/LuLWM15} develop a joint feature learning approach to automatically learn hierarchical representation from raw pixels for face recognition. Figure~\ref{fig:TIFS2015} illustrates the basic idea of the proposed method. First, each face image is divided into several non-overlapping regions and feature weighting matrices are jointly learned. Then, the learned features in each region are pooled and represented as local histogram feature descriptors. Lastly, these local features are combined and concatenated into a longer feature vector for face representation. Moreover, the joint learning model is stacked into a deep architecture exploiting hierarchical information. As a result, the effectiveness of the proposed approach is demonstrated on five widely used face datasets.

Face recognition has been widely used in security systems and human-machine interaction systems. It is still a challenge for computer to automatically identify or verify a person due to large variations, \eg illumination, pose and expression. Deep learning can utilize big data for training deep architecture models so as to obtain more powerful features for representing faces. In future, face recognition systems in smart cities will largely rely on hierarchical features learned from deep models.

\section{Image Classification}
Image classification has been an active research topic in the past few decades. Many methods have been proposed to achieve very promising results by using Bag-of-Words (BoW) representation \cite{csurka2004visual}, spatial pyramid matching \cite{DBLP:conf/cvpr/LazebnikSP06}, topic models \cite{DBLP:conf/cvpr/LiPT05}, part-based models \cite{DBLP:conf/cvpr/FergusPZ03} and sparse coding \cite{DBLP:conf/cvpr/YangYGH09}, etc. However, these methods utilize raw pixel values or hand-crafted features which cannot capture data-driven representations for specific input data. Recently, deep learning has achieved very promising results in image classification. We review some representative works as follows.

\begin{figure}
	\begin{center}
		\includegraphics[width=1.0\linewidth]{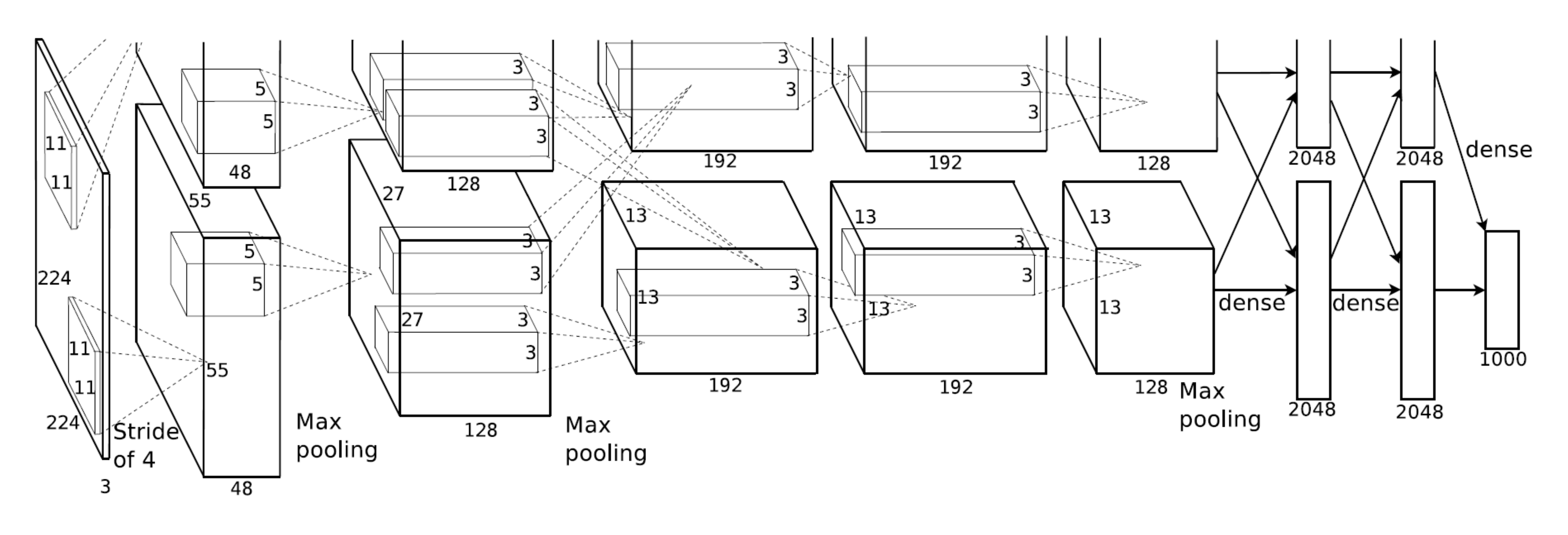}
	\end{center}
	\caption{Illustration of the architecture of the deep CNN proposed by Krizhevsky \etal \cite{DBLP:conf/nips/KrizhevskySH12}. It consists of five convolutional layers, some of which are followed by max-pooling layers, and three fully connected layers with a $1000$-way softmax. }
	\label{fig:CNN2012ImageNet}
\end{figure}

Krizhevsky \etal \cite{DBLP:conf/nips/KrizhevskySH12} develop a deep convolutional neural network which has $60$ million parameters and $650,000$ neurons. The deep model includes five convolutional layers followed by max-pooling layers, and three fully-connected layers with a final $1000$-way softmax. The architecture of the deep convolutional neural network proposed by Krizhevsky \etal \cite{DBLP:conf/nips/KrizhevskySH12} is illustrated in Figure~\ref{fig:CNN2012ImageNet}. The deep model achieved very impressive results on the ImageNet LSVRC-$2010$ contest by reducing the error rates down to $8$\% against the then-state-of-the-art. The dataset includes $1.2$ million high-resolution images in $1000$ different classes. This competition has become one of the largest and most challenging computer vision challenges and is held annually. The challenge has attracted not only academic research groups but also industry companies. For example, Google has won the classification challenge in the ImageNet $2014$ with a error rate $6.66$\%. Nowadays, high performance computing plays a very important role in deep learning. Recently, the Chinese search engine company Baidu has obtained a $5.98$\% error rate on the ImageNet classification dataset by using a supercomputer called Minwa, which consists of $36$ server nodes, each with $4$ Nvidia Tesla K$40$m GPUs. Baidu also claims that Minwa can handle higher-resolution images and the larger training dataset (~$2$ billion images) which is generated by distorting, flipping and changing colors of the original $1.2$ million images. As a result, the system is able to work on real-world photos.

\begin{figure}
	\begin{center}
		\includegraphics[width=1.0\linewidth]{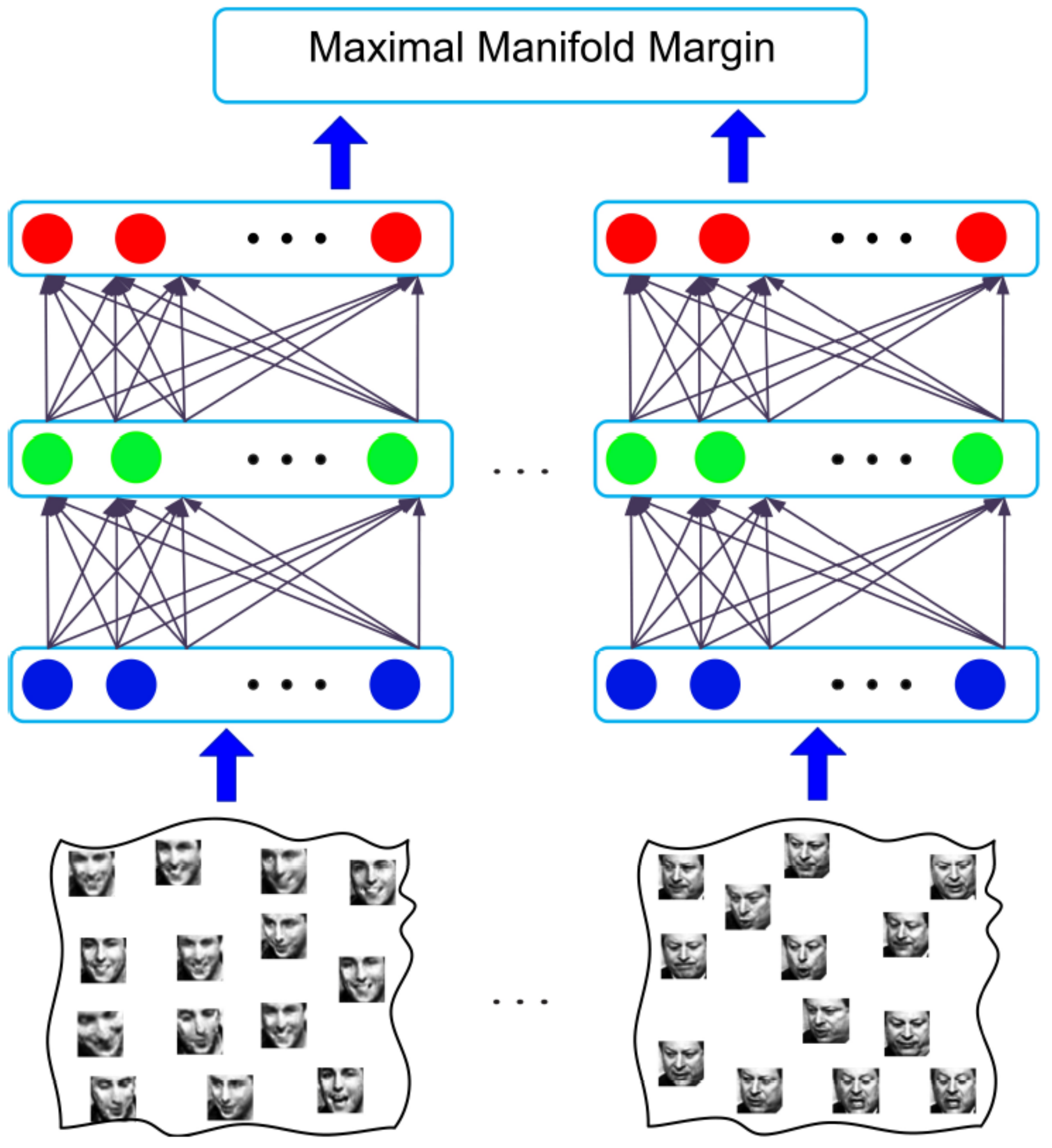}
	\end{center}
	\caption{Illustration of the basic idea of the method \cite{DBLP:conf/cvpr/LuWDMZ15}.}
	\label{fig:LuCVPR2015}
\end{figure}

Lu \etal \cite{DBLP:conf/cvpr/LuWDMZ15} present a multi-manifold deep metric learning (MMDML) method for image set classification. First, each image set is modeled as a manifold which is passed into multiple layers of deep neural networks and mapped into another feature space. Specifically, the deep network is class-specific so that different classes have different parameters in their networks. Then, the maximal manifold margin criterion is used to learn the parameters of these manifold. In the testing stage, these class-specific deep networks are applied to compute the similarity between the testing image set and all training classes. Finally, the smallest distance is used for classification. As a result, the proposed method can achieve the state-of-the-art performance on five widely used datasets by exploiting both discriminative and class-specific information.

\begin{figure}
	\begin{center}
		\includegraphics[width=1.0\linewidth]{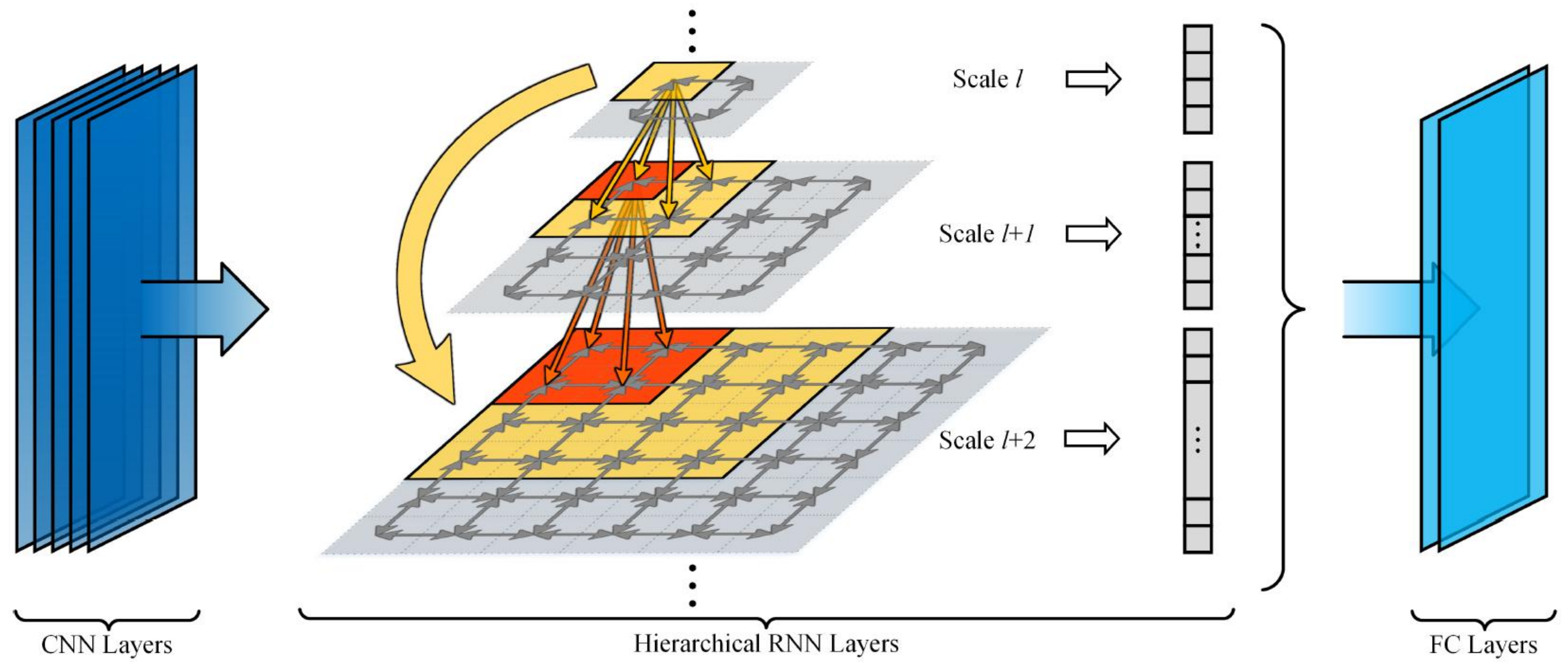}
	\end{center}
	\caption{Illustration of the overall framework of C-HRNNs \cite{DBLP:journals/corr/ZuoSWLWW15}.}
	\label{fig:zuoArxiv}
\end{figure}

Zuo \etal \cite{DBLP:journals/corr/ZuoSWLWW15} develop an end-to-end Convolutional Hierarchical Recurrent Neural Networks (C-HRNNs) to explore contextual dependencies for image classification. Figure~\ref{fig:zuoArxiv} illustrates the overall framework of C-HRNNs. First, mid-level representations for image regions are extracted by using five layers of Convolutional Neural Networks (CNNs). Then, the output of the fifth CNN layer is pooled into multiple scales. For each scale, spatial dependencies are captured by direct or indirect connections between each region and its surrounding neighbors. For different scales, scale dependencies are encoded by transferring information from the higher level scales to corresponding areas at the lower level scales. Finally, different scale HRNN outputs are collected and put into two fully connected layers. C-HRNNs not only make use of the representation power of CNNs, but also efficiently encodes spatial and scale dependencies among different image regions. As a result, the proposed model achieves state-of-the-art performance on four challenging image classification benchmarks.

\section{Scene Labeling}
Scene labeling aims to assign one of a set of semantic labels to each pixel of a scene image. It is very challenging due to the fact that some classes may be indistinguishable in a close-up view. Generally, ``thing" pixels (cars, person, etc) in real world images can be quite different due to their scale, illumination and pose variation. Recently, deep learning based methods have achieved very promising results for scene labeling. We review some of them as follows.

\begin{figure}
	\begin{center}
		\includegraphics[width=1.0\linewidth]{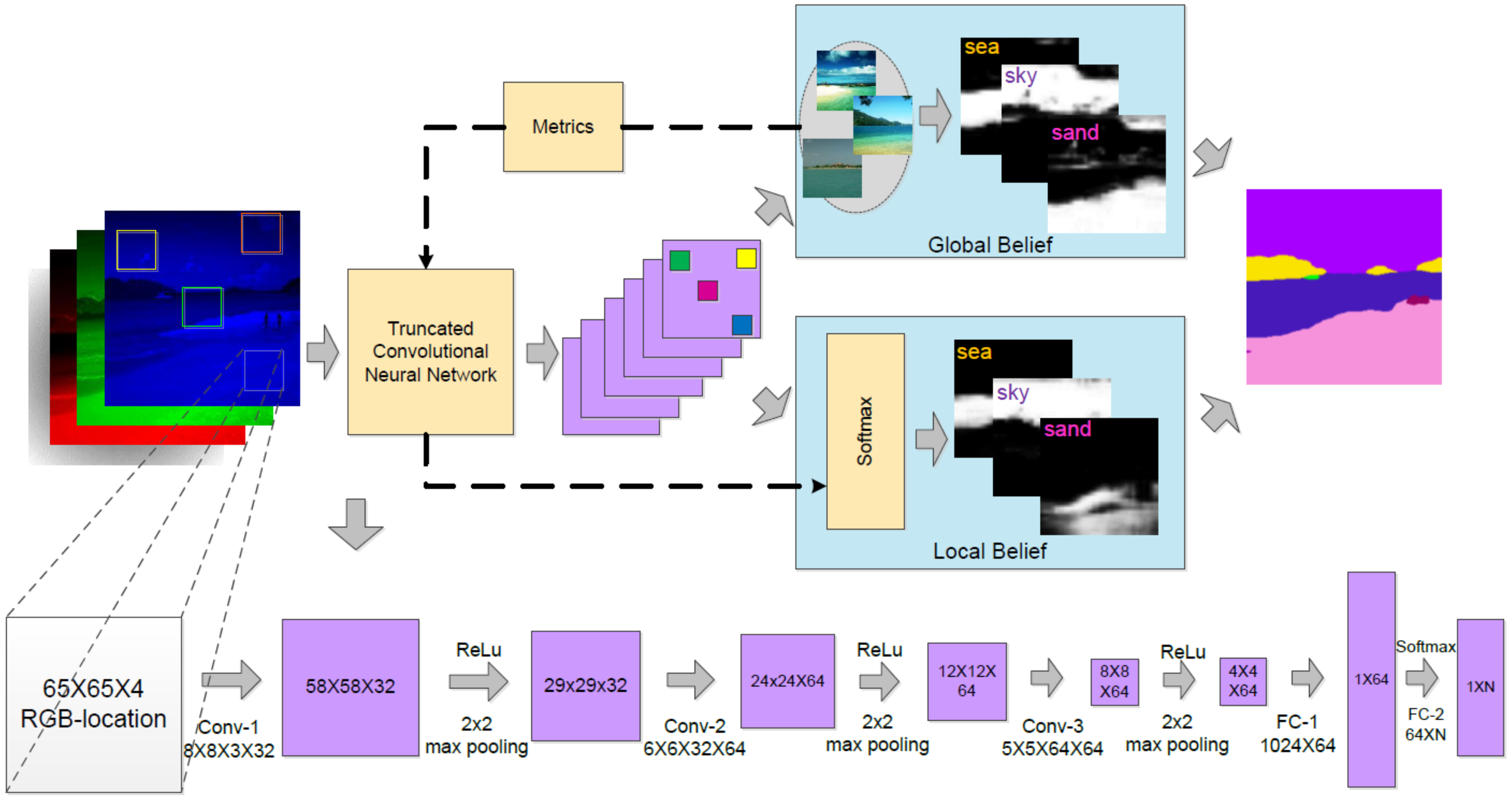}
	\end{center}
	\caption{Illustration of the framework of the approach \cite{DBLP:conf/cvpr/ShuaiWZWZ15}.}
	\label{fig:shuaiCVPR}
\end{figure}

Shuai \etal \cite{DBLP:conf/cvpr/ShuaiWZWZ15} propose to adopt Convolutional Neural Networks (CNNs) as a parametric model to learn discriminative features and classifier for scene labeling. Figure~\ref{fig:shuaiCVPR} illustrates the framework of the proposed method. First, global scene semantics are used to remove the ambiguity of local context by transferring class dependencies and priors from similar exemplars. Then, the global potential is decoupled to the aggregation of global beliefs over pixels. The labeling result can be obtained by integrating the local and global beliefs. Finally, a large margin based metric learning is introduced to make the estimation of global belief more accurate. As a result, the proposed model is able to achieve state-of-the-art results on the SiftFlow benchmark and very competitive results on the Stanford Background dataset.

\begin{figure}
	\begin{center}
		\includegraphics[width=1.0\linewidth]{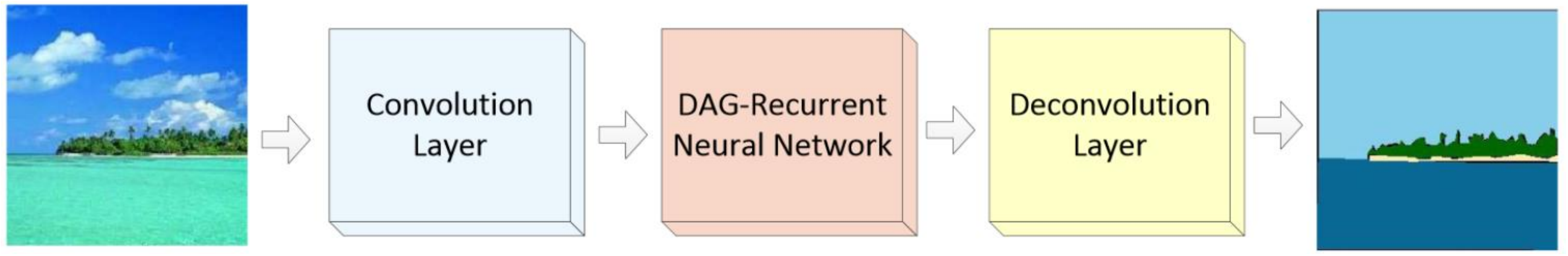}
	\end{center}
	\caption{Illustration of the architecture of the full labeling network in the approach \cite{DBLP:journals/corr/ShuaiZWW15}.}
	\label{fig:shuaiArxiv}
\end{figure}

Shuai \etal \cite{DBLP:journals/corr/ShuaiZWW15} present a directed acyclic graph RNNs (DAG-RNNs) for scene labeling to model long-range semantic dependencies among image units. Figure~\ref{fig:shuaiArxiv} illustrates the architecture of the full labeling network in the proposed method. First, undirected cyclic graphs (UCG) are adopted to model interactions among image units. Due to the loopy property of UCGs, RNNs are not directly applicable to UCG-structured images. Thus, the UCG is decomposed to several directed acyclic graphs (DAGs). Then, each hidden layer is generated independently through applying DAG-RNNs to the corresponding DAG-structured image, and they are integrated to produce the context-aware feature maps. As a result, the local representations are able to embed the abstract gist of the image, so their discriminative power are enhanced remarkably. It is reported that the DAG-RNNs achieve new state-of-the-art results on the challenging SiftFlow, CamVid and Barcelona benchmarks. 

\begin{figure}
	\begin{center}
		\includegraphics[width=1.0\linewidth]{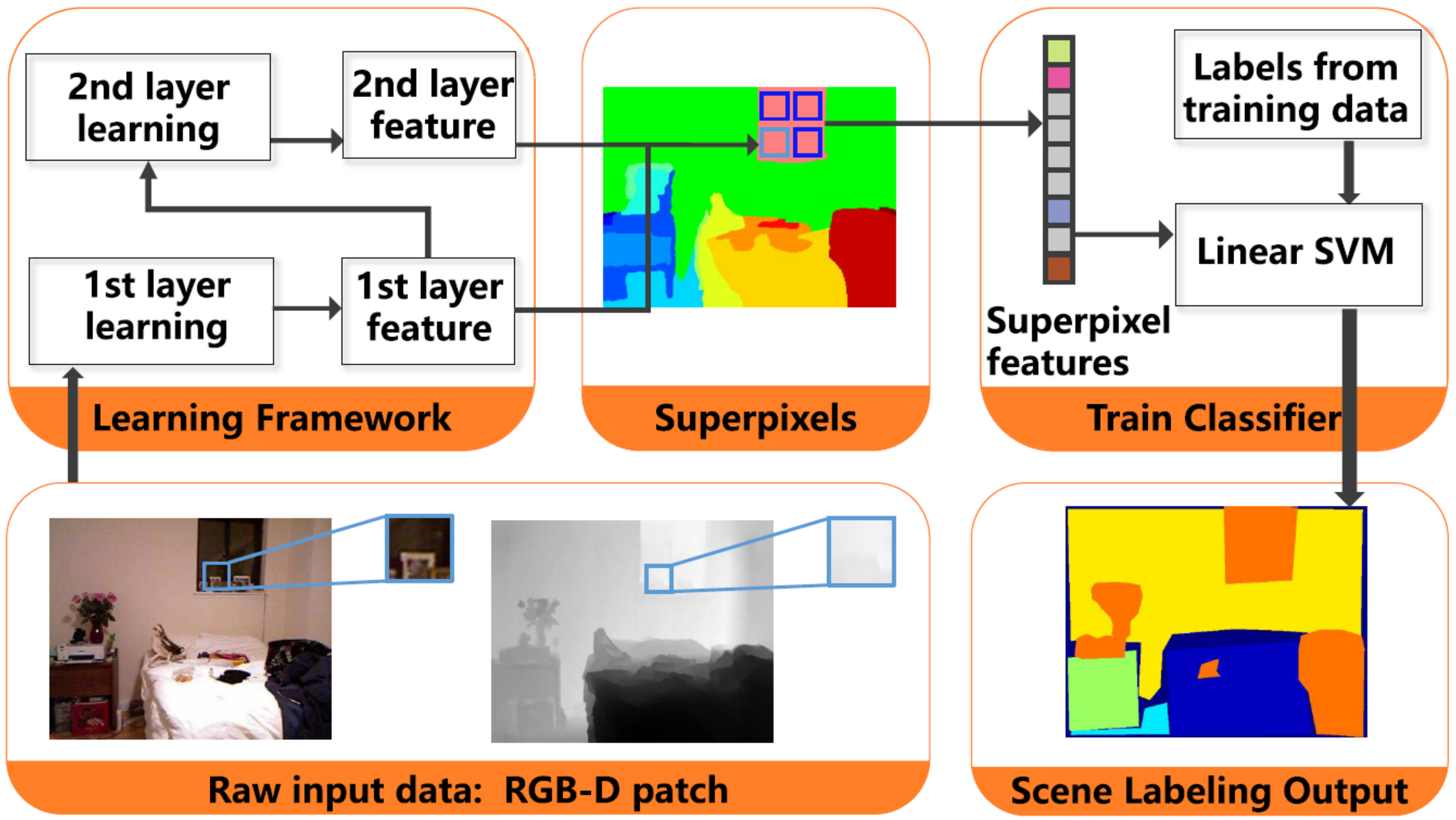}
	\end{center}
	\caption{Illustration of the framework for RGB-D indoor scene labeling in the approach \cite{DBLP:journals/tip/0001LCWC15}.}
	\label{fig:wangTIP_RGBD}
\end{figure}

Wang \etal \cite{DBLP:journals/tip/0001LCWC15} develop an unsupervised joint feature learning and encoding (JFLE) framework for RGB-D scene labeling. Figure~\ref{fig:wangTIP_RGBD} illustrates the framework for RGB-D indoor scene labeling in the proposed method. First, feature learning and encoding are jointly performed in a two-layer stacked structure, called joint feature learning and encoding framework (JFLE). To further extend the JFLE framework to a more general framework called joint deep feature learning and encoding (JDFLE), a deep model is used with stacked nonlinear layers to model the input data. The input to the learning structure (either JFLE or JDFLE) is a set of patches densely sampled from RGB-D images, and the learning output is the set of corresponding path features, which are then combined to generate superpixel features. Finally, linear SVMs are trained to map superpixel features to scene labels. It is reported that the proposed feature learning framework can achieve competitive performance on the benchmark NYU depth dataset.

Scene labeling can be used to understand urban images captured from surveillance cameras. It is a very important component of smart city, in which city elements such as ``road" and ``building" are required to be recognized. Deep models can leverage on big data from smart city to learn hierarchical features for labeling each pixel of scene images. 

\section{Conclusions}
In this paper, we have reviewed the recent progress of deep learning in object detection, object tracking, face recognition, image classification and scene labeling. The deep models have significantly improved the performance in these areas, often approaching human capabilities. The reasons for this success are two-folded. First, big training data are becoming increasingly available (\eg data streams from a multitude of smart city sensors) for building up large deep neural networks. Second, new advanced hardware (e.g. GPU) has largely reduced the training time for deep networks. We believe that deep learning will have a more prospective future in a wide range of applications.

\section*{Acknowledgment}
This research was carried out at the Rapid-Rich Object Search (ROSE) Lab at the Nanyang Technological University, Singapore.  The ROSE Lab is supported by the National Research Foundation, Prime Minister’s Office, Singapore, under its IDM Futures Funding Initiative and administered by the Interactive and Digital Media Programme Office.

\ifCLASSOPTIONcaptionsoff
  \newpage
\fi

% references
\bibliographystyle{IEEEtran}
\bibliography{ref}

%\begin{IEEEbiography}{}
%\end{IEEEbiography}

\end{document}